%% file: seq2seq__MTL_intent_content_extraction_domain_specific_interpreter.tex
\documentclass[twocolumn]{article}
\usepackage[top=2cm, bottom=2.5cm, left=2cm, right=2cm]{geometry}
\usepackage{times,textcomp}
\usepackage[T1]{fontenc}
\usepackage[utf8]{inputenc}
\usepackage{amssymb,amsmath}
\usepackage{enumitem}
\usepackage{eurosym}

\usepackage{titlesec}

\usepackage{titlesec} 
\titleformat{\section}{\large\bfseries}{\thesection}{1em}{\MakeUppercase} 
\titleformat{\subsection}{\large\bfseries}{\thesubsection}{1em}{}
\titlespacing*{\section}{0pt}{\parskip}{0pt}
\titlespacing*{\subsection}{0pt}{\parskip}{0pt}

\usepackage{longtable,booktabs}
\usepackage{graphicx,grffile}
\usepackage{fancyhdr}

\title{\huge\bfseries Seq2Seq and Multi-Task Learning for joint intent and content extraction for domain specific interpreters}

\usepackage{authblk}

\author{Marc Velay}
\author{Fabrice Daniel}

\affil{\small Artificial Intelligence Department of Lusis, Paris, France\\fabrice.daniel@lusis.fr\\http://www.lusis.fr}

\date{July 2018}
\begin{document}

\maketitle

\begin{abstract}
This study evaluates the performances of an LSTM network for detecting and extracting the intent and content of commands for a financial chatbot. It presents two techniques, sequence to sequence learning and Multi-Task Learning, which might improve on the previous task. 
\end{abstract}


\input{chapters/introduction}
\input{chapters/motivations}
\input{chapters/approaches}
\input{chapters/results}
\input{chapters/conclusion}

\newpage

\input{chapters/reference}
\end{document}

%% file: chapters/introduction.tex
\section{Introduction}

Most interfaces today are getting sleeker: decluttering buttons and removing text. In this context, it is positive to substitute large parts of a website with a system to query specific information. This fuelled the chatbot revolution we have witnessed for the part years. Users can now ask a conversational agent their questions. But sometimes, a full conversational chatbot is not the answer, as clients are not looking for a chat but for quick information. The solution is an interpreter, which has been popular in many command line interfaces and modern IDE software. Interpreters are tasked with understanding commands and executing corresponding tasks.

Present solutions including Watson from IBM, DialogFlow from Alphabet or recast.ai were made in answer to the surge in demand for bots. But they were created around a general conversation theme: they should be able to talk about the weather, news and make jokes. They must also be trainable in any subject clients need them. Food companies do not have the same demands as weather services. Some of these systems do not require technical knowledge. This means a company willing to build a chatbot using those technologies will need to gather a dataset and build answers step by step. The final result will not be as efficient as a specialized solution, which would rely on the same dataset and answer-building process. Therefore, for specific fields, specialized interpreters are a better fit. This paper compares different implementations for domain-specific interpreters. 

Chatbots and interpreters depend on the same field of Natural Language Processing. NLP consists of the study of extracting information for machines from text. These information include Intent and Word-Tagging, respectively the desired action and the context of the request. These two information are closely linked, as we can extrapolate an intent using only a tagged sentence, and a list of expected tags from an intent. Current state-of-the-art solutions in NLP are based around recurrent neural networks, such as an LSTM. These are well suited for processing sequences of data, including texts. Being the current best solution, we want to discover the technique best suited for building an interpreter. Presently, Machine Translation is being revolutionized by a new model: sequence to sequence (seq2seq) LSTMs. Interpreters and Machine Translation are related, although they do not have the same objectives. We will also be comparing the efficiency of single-task and multi-task learning. Multi-task Learning is when a single neural network is used to predict several different values, using the same input. It is believed to be more efficient than the traditional approach of single-task networks as learning linked tasks could help learn each task faster \cite{JIDSF}. 

We will start by detailing the information our model can extract from commands in order to process them and how to augment the data. We will then detail the techniques compared in this paper, as well as the chosen encoding and vectorisation techniques. Finally, we will present the results we have achieved. 

%% file: chapters/motivations.tex
\section{Motivations}

Sentences contain several types of information, which we subconsciously understand when talking with someone, but which machines must knowingly extract. These information are intents and context. The intent is what the user wants to do, the context are the elements related to the intent. This is very challenging to do, as a computer only sees a sequence of unicode values, while we see sentences made up of words. Bringing a computer’s understanding to an abstraction close to ours requires teaching it the concept of sentences and the role of words.

For general chatbots, an intent could be to answer a question, to make a joke or to get weather data. In the case of this experiment, based on a finance interpreter, it would be to open new charts, adding indicators or filtering a news feed. An intent is defined using a typical structure in a sentence. There exists only a limited amount of ways to ask someone if it could rain today. When speaking we expect a request to at least have several categories of words. 

These categories make up the context of a sentence. Extracting them is named Part of Speech Tagging, or PoS, in Natural Language Processing. The technique relies on isolating and classifying each word. Classic implementations rely on PoS to extract locations, nouns and verbs. In our application, we want to isolate tickers, indicators and company names. 

Intent and context are closely related. We can extract an intent using only word categories, and generate a list of expected categories from an intent. Therefore, trying to predict one using a deep learning model is linked to predicting the other. Applying Multi-Task Learning is a form of transfer learning: knowledge from one task is used to help understand another task \cite{transferLearning}. 

What we want is to measure the efficiency of MTL compared to single task learning. We also want to quantify the efficiency of sequence to sequence recurrent networks compared to the n-to-m mapping of classic networks. Sequence to sequence models output values until they predict a specific token, known as an end token. That means we do not know the length of the output until the prediction is finished. This type of model relies on its internal state and a feedback loop of its own predictions. 

We hope to find that an MTL sequence to sequence model will be able to generate more accurate results, while being able to generalize better.

%% file: chapters/approaches.tex
\section{Approaches}

There are often several solutions to a problem. We will define every solutions we have tested for each of the components of the interpreter. The first step was to create an algorithm in order to augment the data. Then we had to vectorise the resulting commands. The sequences were processed by machine learning models reliant on different architectures. We will then present the different ways we tested the results. 

Data augmentation is the act of generating new data, in order to artificially increase the quantity of information available to train on. This is necessary as most deep learning algorithms require a large amount of data to train on. Therefore increasing the size of a dataset using data augmentation improves a model’s accuracy \cite{DataAugm}, as demonstrated by previous research.

The first step in created augmented data is gathering actual data. To do this, you must gather samples of what users might say in the selected field. For a financial application, that would be filtering news, manipulating charts by adding indicators or modifying the tracked instruments or even placing trades. We achieve this by brainstorming every question that relates to the application. We find different ways to express the same requests, by varying the order of words, using synonyms or ellipsing words. The gathering of a large sample of sentences is necessary and better results are achieved by soliciting several people for help. 

Once we have gathered a large enough set of samples, we must find internal categories, relations and structures. These properties are used to create a model of the data. The data model will be used to augment the data to train our algorithms on. The result of the previously mentioned analysis of man-made data are those categories. In finance, you need to separate tickers, indicators, graph filters and various numeric values. There are also different actions that need to be discerned, which we defined previously as Intents. It is necessary to separate those in order to generate new sentences procedurally. 

Augmentation is the process of creating new data using information derived from internal and external sources. This is a crucial part for training most Deep Learning algorithms due to the large data requirements. The process also results in more generalized training data, due to creating examples which did not exist beforehand. In signal processing, augmenting data corresponds to creating signals with similar fundamental properties, to which we add gaussian noise. In image processing, applying various filters can have the same effect, such as color shifting, rotations or adding new objects into the image. The new datasets generally result in better metrics, when the model is evaluated.

To apply data augmentation to Natural Language Processing we follow a specific process in order to preserve some rules inherent to languages. When every part of what makes up a sentence and their underlying model has been identified, we can synthesize new sentences. To do so, we separate words by categories, with words that have the same role. We then loop through these categories in order to create many new sentences. For example, when requesting a new chart to be opened “Open EURUSD”, “Open new chart EURUSD”. 

Finally, in order to inflate the generalisation capacities of the model, we can introduce variations in spelling into the words, as well as words without categories. We can introduce these variations by randomly selecting points in a sentence into which we can introduce the variations. We could select the middle of a word and swap two letters to introduce typography mistakes or remove letters to imitate SMS spelling. The random words inserted also act as a new category into which commands which do not relate to finance could be classified. If you ask about the weather, the whole sentence would most likely be classified as not having a category. This was further enhanced by sampling a large corpus for sentences of varying length, in order to have a subset of classless examples. 

The previous process generated several thousand sentences from a couple hundred requests. This large dataset is still in a form that only humans understand. We need to vectorise the sentences in order to apply mathematical operations to them. 

There are several techniques in order to vectorise sentences. The most frequent is word encoding. There are several algorithms for this: Each word is mapped to a corresponding one-hot vector, or to a continuous vector. In the case of one-hot vectors, the system works on a dictionary basis, where each known word equals an entry, and the vector is the length of the known-word dictionary. This system relies on very strict boundaries and learning new words require retraining the whole underlying structure. No knowledge from the words is transferred into the encoding. A continuous vector encoding relies on neighboring words to generate values. The aim is to map the meaning of a word, with similar vectors being neighbors, and a tendency to form cluster by fields. Since each vector holds information about the meaning of a word, it is possible to add new words by retraining only the encoder system and not the solutions relying on them. But it also requires a large domain-specific corpus. There needs to be several occurrences of a word in order to accurately define its meaning. This sort of corpus does not exist for financial information in a shape that would be useful for an interpreter. 

Both styles of word-encoding did not suit our needs. Although these solutions often have better results than letter-encoding \cite{wordEmb}, we found that the later worked very well for our application \cite{letterEncoding}. Therefore we encoded each letter as its equivalent ascii code. The result is a 128-dimension vector. We only went up to 128 and not 256 as most of the remaining characters are unlikely to appear in a keyboard written command. The lower dimensionality usually helps reduce the complexity of a task, making the process of finding a relation between input and output easier. Since each letter is predicted, the output must classify each letter, as it is complicated to predict on a word-level when the input is character-level. That problem would require more computing power to solve compared to categorizing each letter. The output is category for each letter. These are the categories defined previously in the data augmentation process. They include “Buy”, “Instrument” and “Filter”. Each corresponds to one dimension in the one-hot output vector.

We then process these vectors using machine learning algorithms. In our case, we mostly rely on LSTM layers. A Long Short-Term Memory is a recurrent neural network. This means that for each element of a sequence, the network reuses internal states. This has the same effect as human memory, where we remember the context of a previous task while completing a new one. This explains why researchers have started LSTMs and derivates for state-of-the-art applications in many fields. This also explains how the model understands a sequence of letters as a whole sentence. This memory is updated using logical gates, responsible for treating the new data, updating the memory state and the output states. When predicting, the memory state is used to compute the new output state, before both are updated for the next output. 

Usually, a machine learning problem relies on one vector in, one vector out. This is Single Task Learning: you optimize the output function for a single target. Multitask learning is the concept that a neural network could learn several related tasks at the same time. The theory behind this is that we might otherwise overfit a specific aspect of the data while overlooking information that humans would deem important. Training several tasks with shared hidden layers generally helps generalize on the tasks to be learnt.

To achieve multitask learning, there are two techniques: hard or soft parameter sharing \cite{HSPS}. Hard parameter sharing means that shared hidden layers are used. The single dataflow starts at the input and ends when the neural network is split into a specific architecture for each task. The shared hidden layers are then optimized by the sum of the corrections from each loss function. This technique has the lowest chance to overfitting as it is optimized for several different loss functions.

\begin{figure}[h]
\centering
\includegraphics[scale=0.5]{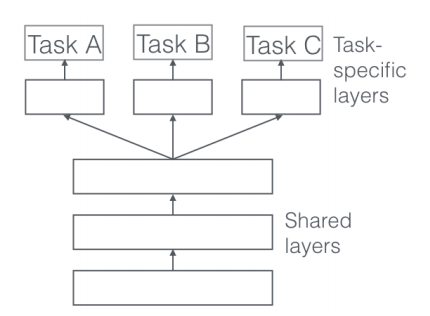}
\caption{{\it Shared layers architecture\/}}
\label{SAL}
\end{figure}

The second technique is soft parameter sharing, where each task has a specific architecture from input to output. Each layer is optimized for the loss function of its specific task, before the parameters are regularized so that they are as close as possible. 

\begin{figure}[ht]
\centering
\includegraphics[scale=0.4]{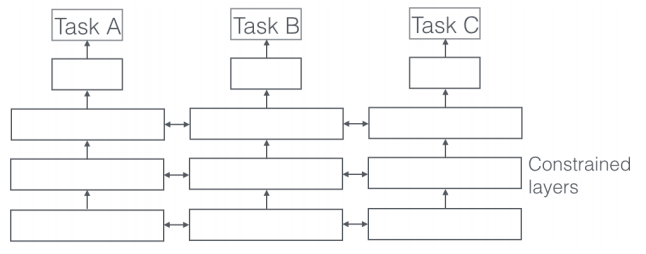}
\caption{{\it Constrained layers architecture\/}}
\label{CAL}
\end{figure}

Sequence to sequence predictions are based on encoding the initial sequence into an intermediary vector and then using it through the decoder to generate the output sequence by iteration. Both models, the encoder and the decoder are based on LSTM architectures. This is due to the ability of the LSTM to embed information in a “memory” state. Therefore, the encoder represents each letter of the sequence inside its inner state. Each sequence results in a unique vector. The encoder relies on a 128-dimensional vector for input, but outputs a 512-dimensional vector. The states are square matrices of 512. We can use this to predict both the intent and tag each letter, using a process akin to information transfer using carrier signals. 

To decode the sequence, the decoder is first initialized with the internal state of the encoder. This vector has all the information from the original sentence embedded in it. We use a “start” token in order to generate the first tag of the sentence. For each successive letter, we feed back the one-hot vector representing the previous tag, as well as the updated state. The decoder generates new tags until the ‘end’ tag is predicted or if we hit the maximum sequence length. Generally, the attention layers make it very improbable to predict sequences longer than the input sequence, in our case.

S2S is trained using the teacher assistance technique. This means that we use one input for the encoder, one input for the decoder and one target for the decoder. We also need to link the initial state of the decoder to the last state of the encoder. We then optimize the model in order to reduce the loss between the decoder output and the target. Each model’s weights are updated accordingly, except the output weights for the encoder. This is due to the actual output of the LSTM being discarded, only the memory state being kept. The resulting vector is the equivalent of a sentence encoding generated from a letter encoding. The encoder input is a sequence of 128-dimensional vectors. The decoder input is a 19-dimensional vector, one for every possible word category. The target for the decoder is the same vector as the input, shifted by one with an end token appended at the end. This is because new predictions depend on previous outputs. 

In the case of multi-target learning, the encoder-decoder architecture remains mostly unchanged. The classifier model, which outputs the intent, relies on the state of the encoder and its output. It is also based on a LSTM architecture, to exploit the embedded letters in memory. We use the output of the encoder as well, due to not having the equivalent of a “start” token for categories.

\begin{figure}[h]
\centering
\includegraphics[scale=0.5]{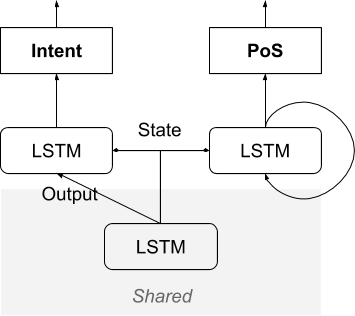}
\caption{{\it S2S MTL architecture\/}}
\label{SML}
\end{figure}

Once the models were trained, we had to test them. Most metrics come from evaluating the models’ loss and accuracy on a validation test. We also tested the models using a more qualitative approach: human testing. The end goal for the application is to be used by humans, so we tested it using a prototype. For the human test, we needed to encode the commands, process them, parse the output, generate answers in human-readable format and modify a mock trading interface. Parsing the predictions means identifying which class the word falls into. We then had to isolate words from specific classes, to be used as keywords in the requests to update the interface. Most functionalities require exact spelling. Therefore we had a list of supported indicators and matched the closest to the user command. This technique was also how we tested if the model generalised well, or not at all. 

%% file: chapters/results.tex
\section{Results}

\begin{table*}[t]
	\centering
	\begin{tabular}{ c c c c c }
		\hline
		{\textbf{Algorithm}} & {\textbf{Intent Accuracy}} & {\textbf{Intent Loss}} & {\textbf{Tagging Accuracy}} & {\textbf{tagging Loss}} \\ \hline
		\textbf{S2S MTL LSTM} & \textbf{0.996} & \textbf{0.006} & \textbf{0.994} & \textbf{0.007}  \\
		MTL LSTM & 0.98 & 0.03 & 0.991 & 0.02\\
		Single LSTM & 0.96 & 0.08 & -- & --\\
		S2S LSTM & -- & -- & 0.97 & 0.02\\ \hline
	\end{tabular}
	\caption{Per algorithm accuracy and loss} 
	\label{tab:results}
\end{table*}

The first tests we did compared single-target networks to multi-task models. We could notice an improvement of 2\% for both tagging and intent detection accuracy. But this difference levels the processing of the interpreter, on known words, to human-like understanding. These last percentages are the key from not being understood to being able to extract information with an acceptable level of error for humans. This increase in accuracy, on a qualitative aspect, has a disproportionate effect on human satisfaction. The MTL model was able to extract information more reliably than the two separate networks. The responses seemed faster. But from manual tests the generalization seemed better on the single intent model. Some sentences which were far in both structure and vocabulary were correctly classified by intent. This might be due to the original data being skewed in favor of a specific intent, corresponding to the successful detections, the result being explained by coincidence more than measurable efforts to create this feature. On these types of sentences, the second model detecting context classified almost every word incorrectly. Therefore the multi-task network classified intent and context more reliably than the two intent and context models. The intent detection algorithm did have a better generalization capacity, although this might be more of a coincidence than an actual functionality, since the tests on other intents had very similar results to the MTL network. 

We then compared the classic end to end detection to the sequence to sequence detection of context. The later model had close to 4\% higher accuracy than the end to end network. The loss was also several factors smaller for the sequence to sequence model. On a qualitative aspect, the seq2seq delimit words much more reliably than end to end, which often did not place spaces at the correct positions. We were also able to encode multi-word indicators into the seq2seq network, being correctly classified as a single entity, instead of being classified as entities of different classes by the end to end model.

Finally, we tested a sequence to sequence multi-task network. This is the most accurate architecture we have conceived. Compared to the classic end to end multi-task model, the accuracy was 1.6\% higher for intent detection and 0.003\% higher for context extraction. Compared to the single purpose end to end models, we saw an increase in intent classification of 3.6\% and context detection of 2.4\%. The quantitative analysis are very similar to the end to end multi-task network compared to single task networks. This network was also the only network capable of correctly classifying sentences which included stock prices and quantities in buy and sell orders. For example, we were able to extract the price, volume and ticker name from “buy 5 @ 295.9 tsla”.

The greatest advantage of the last neural network was the reduction in size and therefore the reduction in computing time. Our final model uses four times less hidden units than any other solution. Comparing the two multi-task networks, the size of the stored parameters went from 230MB to 9MB. This is in addition to a higher accuracy. Both elements are crucial for the user experience, creating a faster and more reliable solution.

%% file: chapters/conclusion.tex
\section{Conclusion}

We have found that a multi-task network, with sequence to sequence learning for content extraction, was the solution with the highest accuracy. This is true for both the loss and accuracy metrics and for a qualitative analysis by humans testers.
	
The main concept lacking from our current solution is generalisation. We can reliably extract intent and context in sentences following the sentence structures we were able to generate. But if we deviate from these structures and vocabulary, the sentences will generally be classified as not having an intent and therefore the context will not be explored. The best solution would be to increase the capacity of our model to generalize. We believe that creating a better algorithm to augment the data would provide such results. 

In order to create a better algorithm, a first step would be to improve the variations added to sentences. At the moment, we only add no-class words to specific parts of the sentences. We would need to also switch more words, find better structures and add spelling errors. These modifications would relate more accurately to a the real use of an interpreter by users. 

The current model is shallow by design, aimed at reducing its overall size and time to compute. Generally, deeper and wider neural networks are known to extract more complex information from the data. The current solution mostly relies on structure and specific word classes being in specific slots. We could hope that a larger network would find more aspects of the commands in order to generalize the future classification.

The current results seem to be high enough to satisfy some users. With some more vocabulary, a business-ready solution can be created. This study was realized in term of a domain specific approach. These techniques most likely would not work for the very large scope of general purpose applications due to the present limits to the capacity of generalizing to new sentences. But the solution for specific domains is fast and efficient, meeting most business-oriented criteria.

%% file: seq2seq__MTL_intent_content_extraction_domain_specific_interpreter.bbl
\begin{thebibliography}{}

\bibitem{JIDSF} 
B. Liu \& I. Lane (2016) \textit{Attention-Based Recurrent Neural Network Models for Joint Intent Detection
and Slot Filling}.

\bibitem{transferLearning} 
J. Huang et al (2013) \textit{Cross-Language knowledge transfer using multilingual deep neural network with shared hidden layers}.

\bibitem{DataAugm} 
J. Wang \& L. Perez (2017) \textit{The Effectiveness of Data Augmentation in Image Classification using Deep
Learning}.

\bibitem{wordEmb} 
JK. Kim et al (2016) \textit{Intent detection using semantically enriched word embeddings}.

\bibitem{letterEncoding} 
D. Golub \& X. He (2016) \textit{Character-Level Question Answering with Attention}.

\bibitem{HSPS} 
S. Ruder (2017) \textit{An Overview of Multi-Task Learning in Deep Neural Networks}.

\end{thebibliography}
